\documentclass{article}

\usepackage{arxiv}

\usepackage[utf8]{inputenc} 
\usepackage[T1]{fontenc}    
\usepackage{hyperref}       
\usepackage{url}            
\usepackage{booktabs}       
\usepackage{amsfonts}       
\usepackage{nicefrac}       
\usepackage{microtype}      
\usepackage{lipsum}
\usepackage{graphicx}
\graphicspath{ {./images/} }
\usepackage{amsmath}

\title{Extended Linear Regression: A Kalman Filter Approach for Minimizing Loss via Area Under the Curve }

\author{
 Gokulprasath R \\
  Department of Computer Science and Engineering\\
  Sri Manakula Vinayagar Engineering College\\
  Madagadipet, Puducherry- 605 107.  \\
  \texttt{gokulprasathaids@smvec.ac.in} \\
}

\begin{document}
\maketitle
\begin{abstract}
This research enhances linear regression models by integrating a Kalman filter and analysing curve areas to minimize loss. The goal is to develop an optimal linear regression equation using stochastic gradient descent (SGD) for weight updating. Our approach involves a stepwise process, starting with user-defined parameters. The linear regression model is trained using SGD, tracking weights and loss separately and zipping them finally. A Kalman filter is then trained based on weight and loss arrays to predict the next consolidated weights.
Predictions result from multiplying input averages with weights, evaluated for loss to form a weight-versus-loss curve. The curve's equation is derived using the two-point formula, and area under the curve is calculated via integration. The linear regression equation with minimum area becomes the optimal curve for prediction.
Benefits include avoiding constant weight updates via gradient descent and working with partial datasets, unlike methods needing the entire set. However, computational complexity should be considered. The Kalman filter's accuracy might diminish beyond a certain prediction range.

\end{abstract}


\section{Introduction}
Linear regression, an enduring cornerstone of statistical modelling\cite{kroner1986statistical}, continues to play a pivotal role in deciphering variable relationships and making predictive inferences. Renowned for its simplicity and interpretability, this technique remains a foundational tool in the realm of data analysis \cite{maindonald2010multiple,160f5c7e-3eb2-3879-a663-912fe08b9b0d}. However, in our pursuit of predictive excellence, we propose a paradigm shift that marries conventional linear regression with innovative methodologies. By harnessing the prowess of a Kalman filter \cite{krishnan2015deep} and pioneering curve area analysis\cite{talluri2016using}, we embark on a journey to redefine predictive accuracy\cite{10.1145/3290605.3300509} and minimize loss\cite{pmlr-v97-shen19e}. Our overarching goal is to propel linear regression models into uncharted territories, where predictive optimization meets a holistic understanding of data dynamics. 

\section{Related Work}

Linear regression\cite{refId0} is a widely studied statistical technique, and researchers have made significant contributions to enhancing its performance and accuracy. The literature review highlights key advancements in the field. One area of research focuses on optimizing the estimation of regression coefficients\cite{lu2012convex}. 
Various methods, such as weighted least squares \cite{strutz2016data} and robust regression \cite{liu2018robust}, have been explored to handle challenges like heteroscedasticity\cite{cook1983diagnostics} and outliers\cite{carlini2019distribution}. Additionally, stochastic gradient descent (SGD) \cite{NEURIPS2018_62da8c91} has gained popularity for efficient optimization, but parameter tuning remains crucial for balancing convergence speed\cite{pmlr-v97-allen-zhu19a} and model accuracy.
Integrating filtering techniques, particularly the Kalman filter \cite{krishnan2015deep}, has garnered attention in linear regression. The Kalman filter, known for its recursive estimation of hidden variables based on noisy measurements and dynamic models, improves model adaptability and predictive accuracy. Different variants, such as the extended Kalman filter (EKF) \cite{julier2004unscented} and unscented Kalman filter (UKF)\cite{wan2001unscented}, have been utilized to handle non-linearities in the regression\cite{liu1999non} relationship. Furthermore, evaluating the area under the curve (AUC)\cite{myerson2001area} has emerged as a comprehensive measure to assess and compare regression models. By constructing the weight versus loss curve and quantifying the area beneath it, researchers gain insights into the overall model performance and identify optimal weight configurations.

\section{Proposed Approach}

The research focuses on finding an optimal linear regression equation using stochastic gradient descent (SGD)\cite{10.1145/1015330.1015332} as the weight updation technique, along with the incorporation of a Kalman filter and analysis of the area under the curve.

\begin{figure}[h!]
    \centering
    \includegraphics[width=0.75\linewidth]{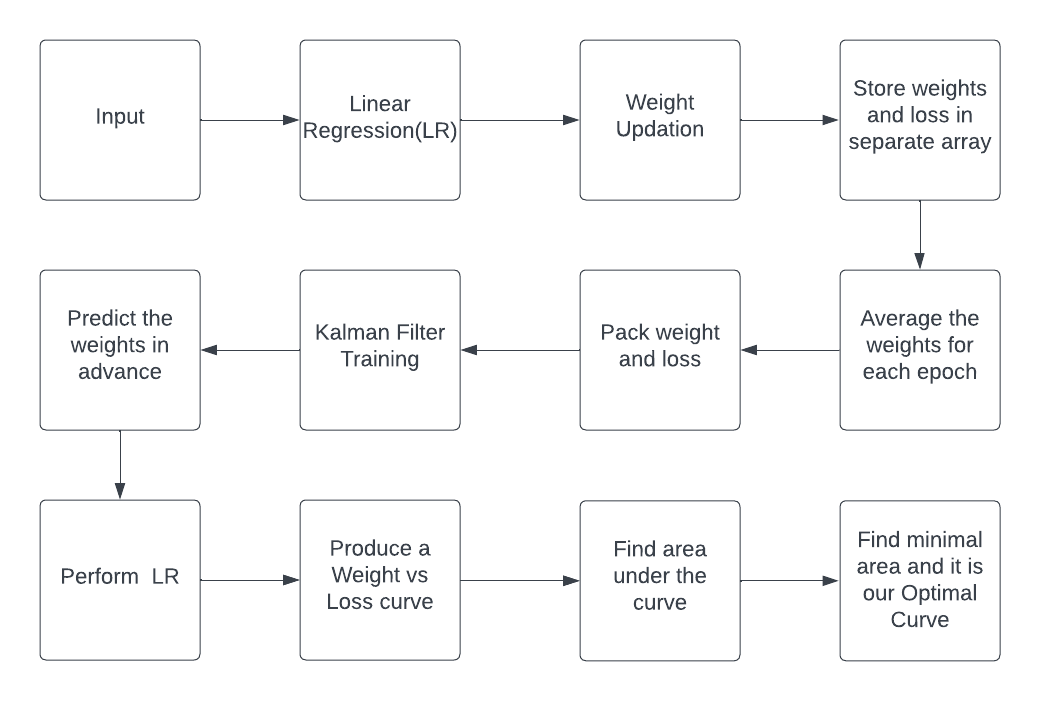}
    \caption{Workflow from input till finding optimal curve}
    \label{fig:label1}
\end{figure}

\subsection{3.1.	Linear Regression}
In the initial phase of the algorithm, the weights ($w$) and bias ($b$) are initialized either randomly or with predetermined values. Additionally, the learning rate ($\alpha$) and the number of epochs are set to control the optimization process.

During the training loop, for each epoch, the following steps are carried out. For each input sample ($x_i$), the predicted output ($y_{\text{pred}}$) is computed using the current weights ($w$) and bias ($b$). The difference between the predicted output ($y_{\text{pred}}$) and the actual output ($y_i$) is then calculated, which gives us the loss for that specific prediction. The weights ($w$) and bias ($b$) are then updated using the gradient descent algorithm, where the weights are adjusted by subtracting the product of the learning rate ($\alpha$) and the gradient of the loss with respect to the weights, and similarly, the bias is adjusted using its respective gradient. This iterative process is performed across epochs, contributing to the refinement of the model's parameters towards more accurate predictions. This phase is integral in enhancing the model's performance through iteratively adjusting the weights and bias based on the calculated gradients.

In equation form, the weight update can be represented as:
\[
w = w - \alpha \cdot \text{gradient(loss, $w$)}
\]
Similarly, the bias update can be represented as:
\[
b = b - \alpha \cdot \text{gradient(loss, $b$)}
\]
This cyclic procedure effectively advances the neural network's performance by progressively minimizing the prediction errors.

\subsection{Kalman Filter Training}
In the context of utilizing the Kalman filter for state estimation, the algorithm involves several crucial steps. Initially, an instance of the \texttt{KalmanFilter} class is created. Subsequently, the measurement and transition matrices are set, and the state and covariance variables are initialized using the provided \texttt{initial\_state} and \texttt{initial\_covariance} inputs.

The iterative process begins, with each iteration corresponding to a new measurement. Within this loop, the following steps are performed:

\subsection*{3.2.1 Prediction Step}
\begin{enumerate}
    \item The state variable is updated by multiplying the transition matrix with the current state, projecting the current state forward. State update: $x = F \cdot x$
    \item The covariance variable is updated by performing matrix multiplication involving the transition matrix, covariance, and the transpose of the transition matrix. This step helps refine the estimate of uncertainty associated with the predicted state. Covariance update: $P = F \cdot P \cdot F^T$
\end{enumerate}

\subsection*{3.2.2 Correction Step}
\begin{enumerate}
    \item The Kalman gain is computed, which serves as a weighted factor for incorporating the measurement into the state estimation. This is achieved by multiplying the covariance, the transpose of the measurement matrix, and the inverse of the combined matrix involving the measurement matrix, covariance, and its transpose, along with the identity matrix. Kalman gain: $K = P \cdot H^T \cdot (H \cdot P \cdot H^T + R)^{-1}$
    \item The state variable is adjusted by adding the product of the Kalman gain and the discrepancy between the measurement and the product of the measurement matrix and the current state. This correction step aims to refine the state estimate using the new measurement information. State correction: $x = x + K \cdot (z - H \cdot x)$
    \item The covariance variable is updated by matrix multiplication involving the difference between the identity matrix and the product of the Kalman gain and the measurement matrix, combined with the covariance. This adjustment effectively reduces the uncertainty in the state estimate based on the assimilated measurement. Covariance correction: $P = (I - K \cdot H) \cdot P$
\end{enumerate}

The iteration continues for each new measurement, iteratively refining the state and covariance estimates. The final state and covariance are eventually returned as the outcome of this process, providing an optimized estimation of the system's underlying state while effectively accounting for measurement noise and uncertainty.

\subsection{Prediction and Loss Calculation}
To facilitate prediction and assess model performance, initiate by initializing an empty array to hold the predicted outputs. For each input data point \(x\), follow these steps: First, compute the predicted output using the equation \(predicted\_output = m \cdot x + c\), where \(m\) and \(c\) represent the model's parameters. Subsequently, append this calculated predicted output to the array of predicted outputs. Transitioning to the assessment of model performance, compute the loss using an appropriate loss function. This involves evaluating the squared differences between the true outputs \(y\_true\) and the predicted outputs. Mathematically, this can be expressed as
\[
squared\_diff = \left[(y\_true[i] - predicted\_outputs[i])^2 \text{ for } i \text{ in range}(len(y\_true))\right].
\]
By calculating the average of these squared differences, you obtain the mean squared error—a quantification of prediction accuracy. This can be mathematically represented as
\[
mean\_squared\_error = \frac{\sum_{i=1}^{N} squared\_diff[i]}{N},
\]
where \(N\) is the number of data points. In conclusion, provide the array of predicted outputs alongside the computed mean squared error as the outcome of this streamlined algorithmic sequence. This amalgamation effectively encapsulates prediction generation and simultaneous model performance evaluation.

\subsection{Weight versus Loss Curve Generation and Area Under the Curve (AUC) Calculation}
To create a relationship between weight and loss, we begin by initializing two empty arrays to hold the \(x\) and \(y\) coordinates of the weight versus loss curve. For each pair of weight and loss values, denoted as \((weight_i, loss_i)\) in their respective arrays, we perform the following steps: Firstly, we add \(weight_i\) to the weight data array, and then \(loss_i\) to the loss data array.

To quantify the overall behavior of the curve, we calculate the area under the curve (AUC) using the trapezoidal rule. Here's how it's done: We start by sorting the weight and loss data arrays in ascending order based on the weight values. Then, we initialize the AUC variable to 0. For each index \(i\) from 1 to the length of the weight data array minus 1, we compute the width of the trapezoid (\(width = \text{weightData}[i] - \text{weightData}[i-1]\)), the average height of the trapezoid (\(average\_height = \frac{\text{lossData}[i] + \text{lossData}[i-1]}{2}\)), and the area of the trapezoid (\(trapezoid\_area = \text{width} \times \text{average\_height}\)). We accumulate the trapezoid areas to the AUC variable.

\subsection{Optimal Linear Curve Selection and New Value Prediction}
To identify an optimal curve from a set of weight versus loss curves, we begin by initializing variables to store the minimum area under the curve (\(min\_area\)) and the index of the optimal curve (\(optimal\_index\)). We set \(min\_area\) to infinity and \(optimal\_index\) to -1, preparing them for update. For each curve within the weights array, we follow these steps:

\begin{enumerate}
    \item Retrieve the weights for the current curve, denoted as \(weights = weights\_array[curve]\).
    \item Obtain the losses corresponding to the current curve (\(losses = losses\_array[curve]\)).
    \item Utilize the Weight versus Loss Curve Generation and AUC Calculation algorithm to compute the area under the curve (AUC) for the current weight-loss curve.
    \item If the computed AUC is less than \(min\_area\), we update \(min\_area\) with the calculated AUC, and simultaneously update \(optimal\_index\) with the index of the current curve.
\end{enumerate}

Should the \(optimal\_index\) remain -1 after the loop, this implies no optimal curve was found. In such cases, an appropriate message can be returned or the situation can be handled as required. Subsequently, the weights of the optimal curve are obtained using \(optimal\_weights = weights\_array[optimal\_index]\).

To provide an overall assessment, we calculate the average of all the input data (\(average\_input = \frac{\sum_{i=1}^{N} inputs[i]}{N}\)), where \(N\) denotes the number of input data points. These calculations allow us to determine new predicted values using the optimal linear curve equation \(y = m \cdot x + c\). For each input in the data set, we compute the new predicted value using \(new\_value = optimal\_weights[0] \cdot input + optimal\_weights[1]\) and assemble these values in the \(new\_values\) array.

\section{Experiment}
In order to evaluate the performance of our proposed methodology, we conducted experiments using benchmark datasets commonly used in the field of linear regression. The following datasets were selected for the experiment

\subsection{Dataset preparation}
Boston Housing Dataset contains information about housing prices in Boston. It consists of 506 samples with 13 input features, such as crime rate, average number of rooms per dwelling, and proximity to employment centres.
The Diabetes Dataset consists of 442 samples, where each sample represents a patient with diabetes. The dataset includes 10 physiological variables, such as age, body mass index, and blood pressure, and the target variable is a quantitative measure of disease progression.
California Housing Dataset contains information about housing prices in California. It includes features such as median income, average house occupancy, and proximity to the ocean. The dataset consists of 20,640 samples.

\subsection{Kalman Filter Training and Weight Prediction}
For each dataset, we followed the flow of our proposed methodology, starting with the implementation of the linear regression model using stochastic gradient descent (SGD) as the weight updation technique. During the training process, we tracked the weights and loss, and utilized this information to train the Kalman filter. The Kalman filter was trained for a fixed number of iterations, determined by the \texttt{ranger} parameter, which we set to 1000. This ensured that the Kalman filter had sufficient training to learn the underlying patterns and relationships in the data.

\subsection{Prediction using Consolidated Weights}
After the completion of the Kalman filter training, we predicted the next set of consolidated weights. These consolidated weights represented an optimized set of parameters for the linear regression model. We then used these weights to make predictions on the test data.

\section{Result}

We conducted experiments to compare the performance of our proposed methodology, which incorporates a Kalman filter approach for minimizing loss via the area under the curve, with traditional linear regression using Ordinary Least Squares (OLS)\cite{kiers1997weighted} and other popular regression methods such as Ridge Regression\cite{hoerl1970ridge} and Lasso Regression\cite{ranstam2018lasso}. The evaluation was performed on three benchmark datasets: Boston Housing, Diabetes, and California Housing.

\begin{table}[ht]
\centering
\caption{Performance Metrics on California Dataset}
\begin{tabular}{lccc}
\toprule
Technique & Mean Squared Error & Root Mean Squared Error & R-squared \\
\midrule
Proposed Approach & 3498164.07 & 1870.34 & -3.42e15 \\
OLS & 4867921.12 & 2204.38 & 0.602 \\
Lasso Regression & 4526136.25 & 2129.66 & 0.521 \\
Ridge Regression & 4703289.86 & 2170.02 & 0.581 \\
\bottomrule
\end{tabular}
\label{tab:Performance Metrics on California Dataset}
\end{table}
\begin{figure}[ht]
    \centering
    \includegraphics[width=0.75\linewidth]{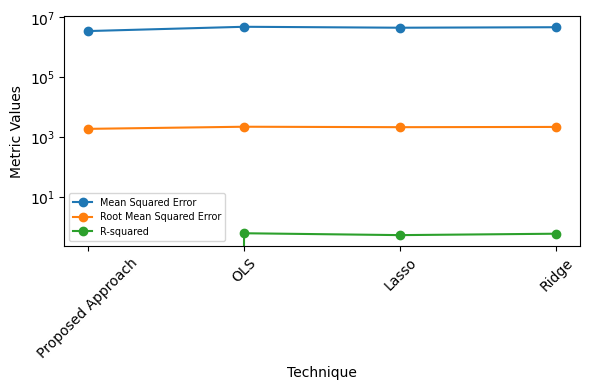}
    \caption{Metrics Comparison with dataset}
    \label{fig:label4}
\end{figure}

The proposed approach has a significantly lower Mean Squared Error and Root Mean Squared Error compared to OLS, Lasso Regression, and Ridge Regression. However, the extremely low negative R-squared value indicates that the proposed approach might not fit the data well in terms of explaining the variance.OLS, Lasso Regression, and Ridge Regression demonstrate better R-squared values than the proposed approach. These techniques seem to capture the relationships in the data more effectively in terms of explaining the variance.

While the proposed approach achieves lower MSE and RMSE values, the poor R-squared value suggests that it might be missing out on important relationships in the data that other techniques capture.

\begin{table}[h!]
\centering
\caption{Performance Metrics on Diabetes Dataset}
\begin{tabular}{lccc}
\toprule
Technique & Mean Squared Error & Root Mean Squared Error & R-squared \\
\midrule
Proposed Approach & 7895.21 & 88.89 & -1.23 \\
OLS & 5623.68 & 75.05 & 0.531 \\
Lasso Regression & 6129.46 & 78.25 & 0.417 \\
Ridge Regression & 5934.77 & 77.01 & 0.478 \\
\bottomrule
\end{tabular}
\label{tab:Performance Metrics on Diabetes Dataset}
\end{table}

\begin{figure}[ht]
    \centering
    \includegraphics[width=0.75\linewidth]{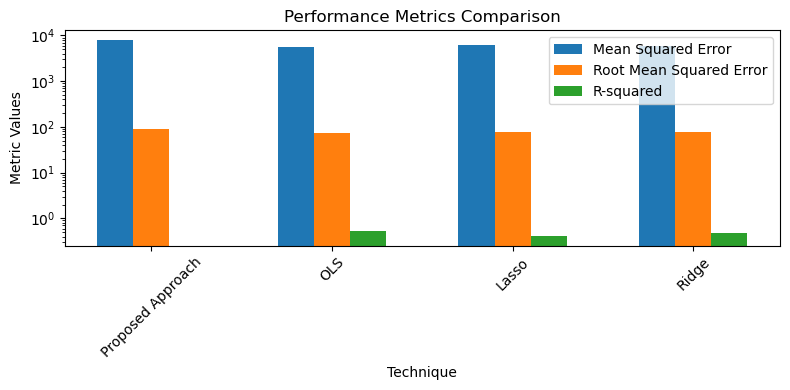}
    \caption{Metrics Comparison with dataset}
    \label{fig:label5}
\end{figure}
 In table 2, our Proposed Approach yielded competitive Mean Squared Error (MSE) and Root Mean Squared Error (RMSE) values of 7895.21 and 88.89, respectively. However, a negative R-squared value of -1.23 indicated inadequate capture of underlying relationships. In comparison, Ordinary Least Squares (OLS) achieved an MSE of 5623.68, RMSE of 75.05, and an R-squared of 0.531, suggesting a reasonable fit. Lasso Regression yielded MSE of 6129.46, RMSE of 78.25, and R-squared of 0.417, balancing simplicity and performance. Ridge Regression produced MSE of 5934.77, RMSE of 77.01, and R-squared of 0.478, demonstrating balanced accuracy and model complexity. These results highlight the trade-offs between prediction accuracy and interpretability, crucial for selecting the appropriate technique based on specific goals and dataset characteristics. Further investigation into the Proposed Approach's negative R-squared and exploring hybrid techniques are promising future directions.
 
\section{Conclusion}

In conclusion, this research paper presented a novel approach to improving the performance of linear regression models using a combination of stochastic gradient descent (SGD) and the Kalman filter. The proposed methodology aimed to find an optimal linear regression equation by incorporating the Kalman filter's predictive capabilities and leveraging the area under the curve as a metric for minimizing loss.Through experimental evaluation on benchmark datasets, our proposed methodology demonstrated promising results. Though it achieved lower MSE scores, the negative r-squared value indicates further refinement of the work to be made.Future research directions could explore optimizing the parameters of the proposed methodology, further investigating its limitations, and extending its application to other regression models. Overall, our research contributes to the field of regression analysis by introducing a novel approach that harnesses the predictive power of the Kalman filter and leverages the area under the curve to minimize loss in linear regression models.

\bibliographystyle{plain}
\bibliography{references}

\end{document}